\documentclass[11pt]{article}

\usepackage[final]{acl}

\usepackage[utf8]{inputenc}
\usepackage[T1]{fontenc}
\usepackage{times}
\usepackage{latexsym}
\usepackage{graphicx}
\usepackage{booktabs}
\usepackage{array}
\usepackage{amsmath,amssymb}
\usepackage{url}
\usepackage{tikz}
\usetikzlibrary{arrows.meta,positioning,calc}
\title{Corpus Quality Improves, Retrieval Utility Does Not:\\
An Ablation of Preprocessing and Refinement in a Speech-to-QA Pipeline}

\author{Wonsup Shin \\
  Flitto \\
  \texttt{wonsup.shin@flitto.com} \\\And
  Jingu Kim \\
  Flitto \\
  \texttt{jin@flitto.com}}

\begin{document}
\maketitle

\begin{abstract}
\noindent
Practitioners building training corpora with a pipeline of language models keep
a preprocessing or refinement (post-process) stage when switching it on raises
the corpus quality. We ablate a production pipeline that turns conference speech
into question--answer pairs. The preprocessing stage rewrites the transcript,
and the refinement stage filters the generated pairs. Switching each on and off
gives four corpora from the same 40 Korean medical and finance sessions. On the corpus itself, quality rises as the stages
switch on. Four LLM judges and six human domain experts rank the four corpora
the same way, and the fully refined corpus leads by $+0.18$ and $+0.22$ on
a 1--5 scale. We then ask whether that gain carries downstream. We supply each
corpus to six models at inference and measure retrieval accuracy. The ranking
does not hold. Enabling the transcript rewrite lowers retrieval accuracy by
$0.061$, the only stage effect that survives correction for multiple
comparisons. That stage also reads the whole session, so it almost
doubles the cost of one. With both stages on, accuracy does not change. An equivalence test we
declared in advance confirms this. The
corpus itself is clearly valuable. It lifts retrieval accuracy from $0.231$ to
$0.748$ in every model and domain we test. More than a third of its pairs carry knowledge
that neither of two frontier models produces on its own. On our data, the null does
not come from weak corpora. A rising score on a rubric of this kind did
not predict what the corpus holds for a model.
\end{abstract}

\section{Introduction}
\label{sec:intro}

Practitioners building training corpora with a pipeline of language models face
a practical question. Such a pipeline has several preprocessing and refinement
stages, each of them costly. One stage in the pipeline we study rewrites the
speech transcript before generation. It reads the whole session rather than
finished question--answer pairs, so turning it on almost doubles the cost of a
session and adds six minutes of wall-clock time. The question is which stages are worth that cost.

The usual answer is to score the corpus. A judge rates a sample before and
after a stage, the score goes up, and the stage stays. Sometimes a human panel
confirms the gain. The procedure is reasonable, and it rests on an assumption
that is rarely tested: a corpus that scores higher works better.

We test that assumption on a production pipeline. It turns conference speech
into question--answer pairs and has two stages that can be switched off, one
preprocessing and one refinement. One rewrites the transcript before generation, and
the other filters generated pairs. Switching each on and off gives four corpora from the same 40 Korean medical and
finance sessions.

\begin{figure*}[t]
\centering
\includegraphics[width=0.92\textwidth]{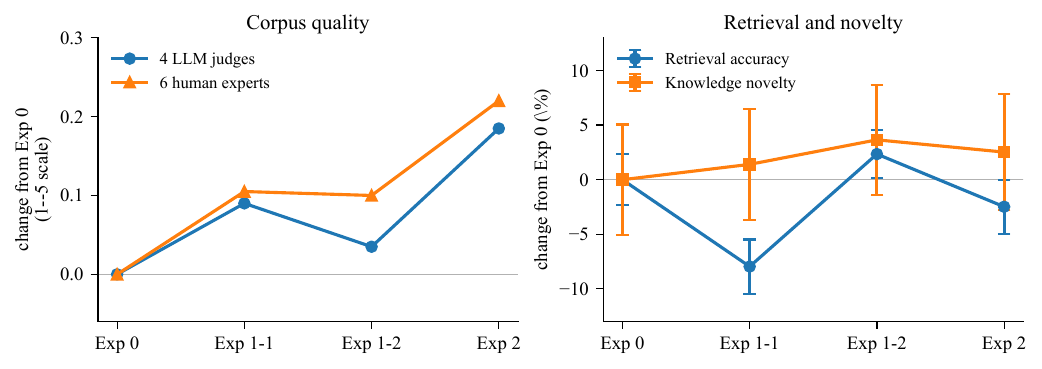}
\caption{The same four corpora, measured two ways. Left: two scoring sources,
each shifted to start at its value for Exp~0, the unrefined corpus, so they
share an axis. Right: two
downstream instruments, in percent change from Exp~0, with 95\% intervals.
Retrieval intervals come from the six models crossed with the two domains
(\S\ref{sec:setup:retrieval}). Novelty intervals come from the pairs. Note the different vertical scales.}
\label{fig:dissociation}
\end{figure*}

On the corpus itself, quality rises as the stages switch on. Four LLM judges and
six human domain experts rank the four corpora the same way. Each stage helps
alone, and both together help most, by $+0.18$ and $+0.22$ on a 1--5 scale. All
five rated dimensions move up (Figure~\ref{fig:dissociation}, left).

That ranking does not carry to retrieval. We supply each corpus to six
instruction-tuned models at inference and measure retrieval accuracy. This needs
no training run, so a corpus that cannot help here is unlikely to help when a
model must learn from it. The fully refined corpus is not the best of the four
(Figure~\ref{fig:dissociation}, right). Each stage moves accuracy the opposite
way. The transcript rewrite lowers
accuracy by $0.061$. The QA filter raises it by $0.018$, which we cannot
separate from noise. Run together, the two stages cancel. The corpus ends
$0.020$ below where it started, and that gap falls inside a bound we declared in
advance.

Part of the reason lies in what the two stages do. The QA filter passes two
thirds of its pairs through unchanged, character for character, and its prompt
forbids new factual claims. The transcript rewrite is more invasive and carries
the penalty, but it rebuilds the corpus rather than editing it, which limits
what we can attribute the penalty to.

The corpus itself has clear value, which sharpens the dissociation. Supplying it
raises accuracy from $0.231$ to $0.748$ in every one of the six models on both
domains, and the gain over a wrong-domain control is $+0.648$. We also ask how much of each
corpus a frontier model already knows. More than a third of the pairs carry knowledge that neither of two frontier
models produces on its own, and the four corpora do not differ
(Figure~\ref{fig:dissociation}, right). The corpus is valuable.
Refining it does not make it more valuable.

We make three contributions. First, we show a dissociation between corpus quality scores and what the corpus
holds for a model, with human experts on the quality side and two instruments on
the other. Earlier studies report that fine-tuning is a weak route for
knowledge injection \citep{zhou2023lima,ovadia2024finetuning,gekhman2024newknowledge}. Our result differs in kind: the corpus improved by every
measure practitioners would use, and neither instrument moved with it.
Second, neither stage adds knowledge. The share of pairs a frontier model
cannot answer is flat across the four corpora, and the stages that raise the
quality score move retrieval slightly down. This suggests they improve how the corpus reads,
not what it holds. The full pipeline is therefore not always the right
choice. When the goal is to extract knowledge, our results give no reason to pay
for either stage: leaving both off costs less than half as much and gives a corpus that
retrieves as well and is no less novel.
Third, we propose a retrieval protocol for judging the QA a speech pipeline
produces. The corpus goes to a model at inference, and closed-book questions
bound how much a model does not already know. The control is a corpus from the
same pipeline with only the domain wrong. Supplying it drops accuracy to
$0.100$, below the $0.231$ a model reaches with no corpus at all, so a gain over
that control cannot come from merely having text to read. The protocol also needs a
wording filter. Questions taken from the transcripts a corpus was built from leave
their wording in the reference answers, and scoring then favors whichever corpus
shares it. The more such questions we admit, the more the
comparison shifts toward the corpora that share the wording. We see this across
nine settings of the filter. That shift is what the wording alone would produce. We fix its
threshold without reference to any accuracy score.

\section{Related Work}
\label{sec:related}

\paragraph{Building and filtering synthetic training data.}
One line of work constructs training data with large language models, from seed
instructions \citep{wang2023selfinstruct,xu2024wizardlm} or from documents
\citep{benallal2024cosmopedia,li2024glan,yang2025entigraph,yehudai2024genie},
and Nemotron-4 340B constructs almost all of its alignment data this way
\citep{nvidia2024nemotron}.

A second line keeps part of that data and drops the rest, from LIMA's $1{,}000$
hand-picked examples \citep{zhou2023lima} to automatic scoring that keeps a
fraction \citep{chen2024alpagasus,liu2024deita,li2024superfiltering}. All of this work validates a score in one direction: select on the score, then show that the
selected data trains a better model. \citet{xia2024rethinking} replicate several
such methods at scale and find few beat random selection. Our refinement stage is a filter of this kind, and we report what it changes and what it leaves alone.

\paragraph{LLM judges as measures of data quality.}
LLM-as-judge scoring is the common way to measure data quality
\citep{zheng2023judging}. Judges agree with human raters often enough to be
useful and run cheaply at scale. Their scores also follow surface
properties: judges favor their own generations
\citep{panickssery2024selfpreference} and longer answers
\citep{dubois2024length}.

That line of work asks whether a judge score matches human preference. We ask
whether a gain that judges and humans both confirm predicts anything downstream.
Four LLM judges and six human domain experts rank our four corpora the same way,
so the gain is not an artifact of one judge, yet it does not carry to retrieval
accuracy or knowledge novelty. That gap is our main result.

\paragraph{Knowledge injection.}
Where new knowledge can enter a model is a settled question in prior work. Fine-tuning
teaches format and style while knowledge comes from pretraining
\citep{zhou2023lima}. Retrieval beats fine-tuning on new facts
\citep{ovadia2024finetuning}, fine-tuning on unheld facts raises hallucination
\citep{gekhman2024newknowledge}, and a model extracts a fact only after seeing
it in varied form \citep{allenzhu2024physics}. Every training route is lossy for
knowledge, so a null under training would confound the corpus with the route.
Retrieval avoids that by supplying the content to the model at inference
\citep{lewis2020rag}, which makes it the most favorable and the cheapest test of content
value. It is our main instrument; we use fine-tuning only as a consistency check.

\paragraph{Speech corpora and retrieval-based measurement.}
Speech recognition now feeds text pipelines \citep{radford2022whisper}, and
several works construct training data from it
\citep{pan2024cosmic,chen2025livecc,jia2025podgpt}. These target speech or
multimodal ability. None measures what a transcript-preprocessing stage adds to
the corpus that follows it. Retrieval places a corpus beside a model at
inference time \citep{lewis2020rag}, which makes it a direct test of what the
corpus holds. RAGAs scores a retrieval system and holds the corpus fixed
\citep{es2024ragas}. We invert that setup: we fix the retriever, the models,
and the questions, and swap only the index.

\section{Pipeline and Corpus}
\label{sec:pipeline}

\subsection{What the pipeline does}
\label{sec:pipeline:arch}

The pipeline turns conference speech into question--answer pairs in three phases.
We toggle the first and the third and keep the middle one fixed, which gives a
$2{\times}2$ design (Figure~\ref{fig:pipeline}).

An in-house system transcribes the audio into sentences, and Phase~0 rewrites
that transcript. It corrects the text, but it corrects from evidence rather than
from fluency. The model reads two further transcripts of the same audio,
each from a different commercial automatic speech recognition (ASR) system, and
rewrites each sentence with all three in view. The in-house transcript is what
gets corrected, and the other two serve as evidence. When both diverge from it in
the same way, the prompt treats that as grounds for a change. It also applies official spellings from
a domain knowledge base, which is how proper nouns that ASR misses are
recovered. The prompt
does not ask it to label speakers or drop fillers. It works in small batches at temperature $0$, and a failed batch
leaves the original sentences in place.

Phase~1 reads the transcript and writes QA pairs, each carrying a domain, a
sub-domain, a difficulty label, and a question type. We hold Phase~1 fixed in
every condition, so the ablation measures the two stages rather than generation.

Phase~2 filters the finished pairs, keeping, rewriting, or dropping each one.
Six steps run in order. Two remove near-duplicates by embedding similarity, one scores pairs with
a judge, two drop pairs whose wording is tied to the session, and one rewrites the
pairs the judge rejected.

The judge is a single model that scores each pair on four dimensions. A pair that
clears every dimension passes untouched. One that does not enters a refinement loop
of at most three rounds, and it is exported only if a later round clears the same
cutoffs. Pairs that never clear them are dropped. Appendix~\ref{app:pipeline}
gives the cutoffs. The prompt forbids new factual claims, so
the stage cannot add facts. One of the
four judge dimensions asks whether a pair stands on its own without the session
around it. We return to both in \S\ref{sec:mechanism}.

\textbf{Exp~0} runs neither stage, \textbf{Exp~1-1} runs Phase~0 alone,
\textbf{Exp~1-2} runs Phase~2 alone, and \textbf{Exp~2} runs both. The two
stages act at different points, and this shapes what we can compare. Phase~0
changes the transcript, so Phase~1 regenerates every QA pair from new text:
turning it on yields a different corpus, not an edited one, and only $1.3\%$ of
the pairs in Exp~1-1 carry a question that also appears in Exp~0. Phase~2 acts on finished pairs,
labeling each and rewriting only the ones it rejects: in the two contrasts that
toggle it, $68\%$ and $64\%$ of its input comes out unchanged, character for
character (App.~\ref{app:pipeline}, Table~\ref{tab:phase2}). Pair-level comparison is therefore available for the Phase~2
contrasts but not the Phase~0 ones, and we report corpus-level statistics where
pair matching is not available.

\subsection{Corpus}
\label{sec:pipeline:corpus}

The corpus covers 40 conference sessions, 19 medical and 21 finance, from
speakers who consented to research use. We capture the audio live at the venue with our
own recording stack, so it is clean. The pipeline handles several languages, and we
draw only on Korean sessions. Segments spoken in another language are kept in
that language rather than translated.

Each condition yields between $2{,}598$ and $2{,}765$ pairs, the full output of
the pipeline on these sessions, or $10{,}698$ pairs across the four. The
same question often appears in more than one condition, so the four hold
$6{,}888$ distinct ones. Mean answer
length grows from $160$ characters in Exp~0 to $192$ in Exp~2, mostly from
Phase~2.

\section{Experimental Setup}
\label{sec:setup}

We ask whether the two stages make the four corpora more useful.
One instrument scores the pairs themselves (\S\ref{sec:quality}). Two measure
the corpus against a model's own baseline (\S\ref{sec:retrieval},
\S\ref{sec:novelty}).

\subsection{Corpus quality}
\label{sec:setup:quality}

We draw 200 QA pairs, 25 per condition per domain. Four LLM judges and six human domain experts score each pair from 1
to 5 on five dimensions: faithfulness, domain accuracy, question quality,
answer depth, and coherence. No rater sees which condition produced a pair. Appendix~\ref{app:setup} lists the judges, the
rubric, and the expert protocol.

\subsection{Retrieval utility}
\label{sec:setup:retrieval}

A corpus is useful if a model answers better when it can read it. We test this with 163 questions. A domain expert wrote 43
of them from the session topics. For the other 120, three per session, a
language model read one session transcript and wrote a question with its
reference answer. Neither writer saw any of the four corpora. That is not
enough. The transcript a writer reads leaves its wording in the reference
answer, and scoring then favors whichever corpus shares it. We call this the
wording effect. A wording filter therefore keeps only the 115 questions whose
reference answer does not lean toward the refined or the unrefined corpus. Its
threshold is fixed on the expert questions without reference to any accuracy score.
Appendix~\ref{app:qsource} gives both values and reports every threshold.

For each question we retrieve the three closest pairs with
\texttt{bge-m3}~\citep{chen2024bgem3}. We run six instruction-tuned open-weight
models from four families, 2.4B to 9B parameters, with greedy decoding
(App.~\ref{app:setup}). We compare eight conditions. Four are the ablation
corpora. One supplies \textbf{no corpus}, and a \textbf{mismatched} control
supplies the other domain's Exp~2 corpus. The last two replace our in-house transcripts with
Whisper-medium output, running Phase~2 in one condition and not the other. They
test whether the results hold when a public system transcribes the speech. Those two
were generated only on the 43 expert-written questions, so their contrasts use
the 33 of those that pass the wording filter. A model may score higher just for
receiving extra text, or because retrieval reshapes its answer format. The
mismatched control supplies text of the same kind from the same pipeline with
only the domain wrong, so a gain over it is a gain from content.

Two judges from different providers score each answer against the reference answer as
correct, partial, or incorrect, mapped to $1.0$, $0.5$, and $0.0$. Judge inputs
carry no identifier of model or condition, and we instruct the judges not to reward
longer answers. A model's scores correlate across questions, so we work at the
level of one model crossed with one domain, which we call a \emph{cell}. Within
a cell we compare conditions question by question and average the differences.
That gives 12 scores, and every interval comes from resampling them.

We release a separate 200-pair sample of the corpora and 10 of these questions
(App.~\ref{app:repro}).

\subsection{Knowledge novelty}
\label{sec:setup:novelty}

Retrieval shows that the corpus helps, not that its knowledge is new. We
therefore ask two frontier models, \texttt{claude-sonnet-5} and
\texttt{gpt-5.6-sol}, each question in the corpus with no documents, and check
whether their answers carry the same core content as our reference answer. We
instruct them to say they do not know rather than guess.

A pair is \emph{novel} when the closed-book answer misses the core content,
\emph{grounded} when our answer is real domain content rather than
transcription noise, and \emph{correct} when it is factually sound. A pair is
\emph{effectively novel} when all three hold. Novelty alone would reward noise, which no closed-book answer reproduces and which therefore always counts as new.

Each model grades the other's closed-book answers, so neither judges its own
output. We report two counts. A pair is novel under the \emph{lenient} count when
either model misses the core content, and under the \emph{strict} count only when
both do.
Appendix~\ref{app:setup} gives the question mapping and the rules for combining
the three axes.

\section{Refinement Improves Corpus Quality}
\label{sec:quality}

\begin{table}[t]
\centering\small
\setlength{\tabcolsep}{4pt}
\begin{tabular}{lrrrr}
\toprule
 & Exp~0 & Exp~1-1 & Exp~1-2 & Exp~2 \\
\midrule
\multicolumn{5}{l}{\emph{corpus quality}, 1--5} \\
4 LLM judges    & 3.81 & 3.90 & 3.84 & \textbf{3.99} \\
6 human experts & 3.36 & 3.46 & 3.46 & \textbf{3.58} \\
\midrule
\multicolumn{5}{l}{\emph{downstream}} \\
Retrieval accuracy & 0.767 & 0.706 & \textbf{0.785} & 0.748 \\
Effectively novel, \% & 35.6 & 36.1 & \textbf{36.9} & 36.5 \\
\bottomrule
\end{tabular}
\caption{The dissociation in one place. Best value in each row is bold.
Novelty is the strict count, pairs that both frontier models miss
(\S\ref{sec:setup:novelty}).}
\label{tab:main}
\end{table}

Both scoring sources rank the four corpora the same way, and Exp~2 is top on
both (Table~\ref{tab:main}). From Exp~0 to Exp~2 the gap is $+0.18$
$[+0.06, +0.32]$ for the LLM judges and $+0.22$ $[+0.07, +0.40]$ for the human
experts. All six experts put Exp~2 above Exp~0. Restricting the panel to \texttt{opus-4.7} and
\texttt{gpt-5.4} gives $+0.23$ $[+0.03, +0.44]$, so the ranking does not depend
on the weaker two (App.~\ref{app:extra}, Table~\ref{tab:quality}).

The ranking holds inside every dimension, and no dimension moves the other way.
Faithfulness and domain accuracy, the two that appear to measure content, each
rise by $+0.17$ (App.~\ref{app:extra}). If the study stopped here, the two stages would look worth their cost.

\section{Retrieval Utility}
\label{sec:retrieval}

\begin{figure}[t]
\centering
\includegraphics[width=\columnwidth]{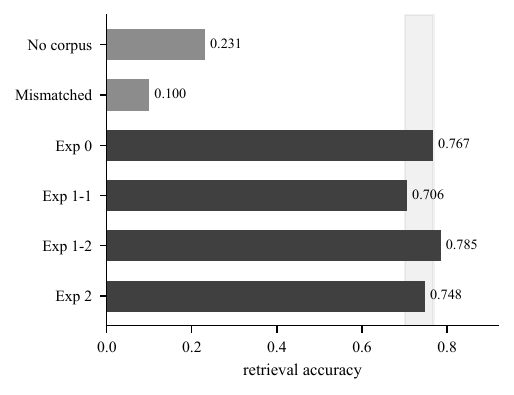}
\caption{Retrieval accuracy with each corpus, on the 115 questions that pass the
wording filter. Grey bars are the two controls, dark bars the four corpora. The
shaded band marks the range the four corpora occupy.}
\label{fig:retrieval}
\end{figure}

\subsection{The corpus adds large value}
\label{sec:retrieval:value}

Accuracy rises from $0.231$ without the corpus to $0.748$ with Exp~2, a gap of
$+0.517$ $[+0.489, +0.544]$ (Figure~\ref{fig:retrieval}). Against the mismatched control the gap is larger,
$+0.648$ $[+0.589, +0.707]$. Both hold in all 12 model-by-domain cells and
survive correction for the nine contrasts we run (Table~\ref{tab:contrasts}).
Wrong-domain passages hurt, dropping
accuracy to $0.100$, below the no-corpus baseline, so the gain is not a side
effect of adding text or of the retrieval format.

\begin{table}[t]
\centering\small
\begin{tabular}{lrr}
\toprule
Contrast & Estimate & $p$ \\
\midrule
\multicolumn{3}{l}{\emph{anchoring}} \\
Exp~2 vs.\ no corpus      & $+0.517$ & $<0.0001^{*}$ \\
Exp~2 vs.\ mismatched     & $+0.648$ & $<0.0001^{*}$ \\
\multicolumn{3}{l}{\emph{ablation}} \\
Exp~1-1 vs.\ Exp~0        & $-0.061$ & $0.0015^{*}$ \\
Exp~1-2 vs.\ Exp~0        & $+0.018$ & $0.025$ \\
Exp~2 vs.\ Exp~0          & $-0.020$ & $0.115$ \\
\multicolumn{3}{l}{\emph{transcript source}, 33 questions} \\
In-house vs.\ Whisper, stages off & $+0.009$ & $0.49$ \\
In-house vs.\ Whisper, stages on  & $-0.001$ & $0.95$ \\
\multicolumn{3}{l}{\emph{novelty, Phase 2}} \\
Exp~1-2 vs.\ Exp~0        & $+1.4$\,pp & $0.30$ \\
Exp~2 vs.\ Exp~1-1        & $+0.4$\,pp & $0.73$ \\
\bottomrule
\end{tabular}
\caption{All nine planned contrasts. An asterisk marks the three that clear
$0.05/9 = 0.0056$. Retrieval estimates are accuracy points on the cell unit;
novelty estimates are percentage points over pairs (App.~\ref{app:stats}).}
\label{tab:contrasts}
\end{table}

\subsection{Running both stages leaves retrieval unchanged}
\label{sec:retrieval:stages}

The ranking of \S\ref{sec:quality} does not appear here. Exp~1-2 is the best of
the four corpora, not Exp~2. The two stages pull in opposite directions. The
transcript rewrite alone drops accuracy from $0.767$ to $0.706$; the QA filter
alone raises it to $0.785$. Table~\ref{tab:contrasts} lists nine planned
contrasts, so we correct for them and require $p < 0.05/9 = 0.0056$. The rewrite
penalty clears that level and the filter gain does not, so we report the
gain as nominal.

Exp~2 and Exp~0 retrieve equally well. Before running the experiment we
registered what would count as a gain worth having, $0.045$, the size the
quality gain would take in retrieval accuracy if it carried one for one
(App.~\ref{app:stats}). The $90\%$ interval for Exp~2 against Exp~0 is $[-0.040,
+0.001]$, which falls inside that bound. The contrast stays inside that bound at
every setting of the wording threshold (App.~\ref{app:qsource}). The
absence of a gain is measured here, not a failure to measure.

Two explanations would undercut this, and neither holds. The first is that
retrieval fails on the refined corpora, so the models never see the right
material. It does not. The top three pairs include one from the question's own
session for $95$ to $98\%$ of questions, with no condition differing from Exp~0.
The second is that the stages merely reword the pairs, and the scores follow that
rewording rather than any content. Lexical drift is too weak for that, correlating
with the accuracy change at only $r=0.10$ (App.~\ref{app:confound}).

Set against the corpus itself, none of these differences is large. The four
corpora span $0.079$, while supplying any of them is worth $0.517$.

\section{Knowledge Novelty}
\label{sec:novelty}

Asked without documents, the two frontier models miss the core content of our
answer at similar rates, $44.3$ to $46.0\%$ of pairs for one and $45.9$ to
$47.3\%$ for the other. Exp~0 is $55.6\%$ effectively novel under the lenient
count and $35.6\%$ under the strict one, so more than one pair in three carries
knowledge that neither frontier model produces on its own.

The four corpora span $1.9$ percentage points on the lenient count and $1.3$ on
the strict one. Every contrast against Exp~0 has a $95\%$ interval that includes
zero, on both counts (Table~\ref{tab:axes}). Pairs cluster inside the 40 shared
sessions, so we read this as small and near zero rather than as a well-powered
null. Appendix~\ref{app:novelty} explains why the measure had little room to move. Phase~2 leaves most pairs untouched, so this contrast is diluted by pairs
that cannot differ. Restricting to the pairs it altered raises the gap but not
out of the noise (App.~\ref{app:nonshared}).

\section{What Each Stage Actually Does}
\label{sec:mechanism}

Two scoring sources show a gain in \S\ref{sec:quality} and two downstream
instruments show none in \S\ref{sec:retrieval} and \S\ref{sec:novelty}. The two
stages do different work, and neither adds much knowledge to the corpus.

Phase~2 keeps two thirds of its pairs unchanged (\S\ref{sec:pipeline}), and among
the pairs it does rewrite, fewer than one in a hundred changes a number or a
term. Its prompt forbids new factual claims, and one judge dimension tests
whether a pair stands on its own rather than what it says, so a pair can keep
every fact, fail on wording alone, and pass once the wording changes. The stage
that raises the quality score moves presentation and leaves content alone.

Phase~0 acts earlier. It edits the transcript, and Phase~1 then writes new pairs
from it, so only $1.3\%$ of the pairs in Exp~1-1 carry a question that also
appears in Exp~0. That
is why pair-level comparison works for the Phase~2 contrasts but not the Phase~0
ones. It also means we cannot say why Phase~0 costs accuracy. Turning it on
changes the transcript and rebuilds every pair at once, and we cannot separate
the two. One candidate is lexical drift, since Phase~0 moves far more text than
Phase~2 does, but \S\ref{sec:retrieval} rules it out at $r=0.10$. What we have
is a pattern, not a cause.

\section{Discussion}
\label{sec:discussion}

A quality score is what keeps these stages in a pipeline. On our corpora it did not predict
what they hold for a model. What to run therefore depends on what the
corpus is for. We measured one pipeline on Korean conference speech with clean
capture, so read what follows as starting points rather than settled defaults.

\paragraph{A corpus people will read.} Refine it. Both stages help and six
domain experts agree.

\paragraph{A corpus for retrieval.} Run the QA filter, which keeps most pairs
and drops weak ones at no retrieval cost, and measure the transcript rewrite in
the target deployment rather than assuming it helps. Where recording quality varies, the rewrite
has more to repair and the balance may tip the other way.

\paragraph{A corpus for injecting knowledge.} Do not expect these stages to
help, since we found no gain on any knowledge axis. The corpus itself more than doubles retrieval accuracy, so more
sessions is the more likely buy. Before
paying for a training run, put the corpus behind a retriever and ask the
questions there. It costs almost nothing, and fine-tuning on these corpora shows
no stage contrast either (App.~\ref{app:secondary}).

\section*{Limitations}

\paragraph{What the equivalence test does and does not settle.} The bound we
registered is tied to our own quality measurement, so practitioners who care
about a smaller gain than we do would need a tighter test.
Pairs also cluster inside the 40 sessions that all conditions share, which a
mixed-effects refit accounts for without changing the estimate
(App.~\ref{app:stats}).

\paragraph{The source audio is already clean.} We capture these sessions live at
the venue with our own recording stack and transcribe them with an ASR model
tuned for it. The audio is clean because the capture is good, not because the
setting is controlled. The transcript entering the pipeline is therefore accurate
to begin with, and the rewrite has little to fix. Our Whisper-medium condition
does not stand in for a noisier setting. Its word error rate against the
in-house transcript runs from $21$ to $37$ percent, and its character error rate
from $8$ to $15$. Korean scoring counts a whole word wrong for a one-character
slip, so the word-level figure overstates how far the two transcripts sit apart.
The corpora built from them are indistinguishable downstream, $+0.009$
$[-0.016, +0.035]$. Noisy or uneven capture is the case our result does not cover, and it is the
first setting we would test next. There the rewrite has real damage to repair,
and the trade-off we report could reverse.

\paragraph{One language, two domains, small models.} Our sessions are Korean
conference speech in medicine and finance. Our six models span $2.4$B to $9$B parameters, the range practitioners serve cheaply, and a larger model may use retrieved text differently.

\paragraph{Continued pretraining is untested.} The pipeline produces QA pairs
for training. Those pairs carry knowledge both frontier models miss
(\S\ref{sec:novelty}), which makes continued pretraining a natural use for them.
We did not test that route, since it costs orders of magnitude more than the
ones we ran. We chose retrieval as the main instrument because it is the most favorable
test of content value, with nothing lost to what training can or cannot absorb.
A corpus that does not help there is unlikely to help through a lossier route,
but the converse does not follow.

\paragraph{Phase~0 ran with two models.} An operational change during data
production switched the rewriting model partway through. Of the
rewriting calls, $77.7\%$ used one model and $22.3\%$ the other. The prompt was identical,
and the switch follows the run date, not the session. The later model covers a
near-identical share of the Phase~0 runs in each domain, $92.1\%$ in medicine
against $92.9\%$ in finance, so the switch adds variance rather than bias.

\section*{Ethics Statement}

The sessions come from service data where speakers consented to research and
non-commercial use. Appendix~\ref{app:repro} lists what we release and the settings behind every
number. The released samples are filtered for personally identifiable
information and for anything that identifies a speaker, an organization, or a
company's own business figures. Human experts were compensated at professional rates and knew how
their ratings would be used.

\bibliography{references}

\appendix

\section{Reproducibility}
\label{app:repro}

We release 200 corpus pairs, 25 per condition per domain, and 10 of the
evaluation questions with their reference answers. The package is at
\url{https://github.com/wonsupshin-flitto/speech-corpus-ablation}. The sessions are third-party
talks, so we keep what reads as domain knowledge on its own and drop anything
that points back at a talk or a company. The same filter runs on all four
conditions and uses no quality signal, so it favors none of them. We also
release the per-item judge verdicts for every instrument, with the maps that
attach each verdict to a model, a domain, a question, and a condition, and the
scripts that produce every number and figure in this paper. Retrieval uses \texttt{bge-m3} with $k=3$ and greedy
decoding at a 512-token limit; judges run at temperature $0$. Bootstrap
intervals use $10{,}000$ resamples with a fixed seed. The code carries an MIT
license and the data a CC BY-NC 4.0 one, which matches the consent the speakers
gave. Appendix~\ref{app:stats} states the analysis choices, and
Appendix~\ref{app:pipeline} gives the pipeline cutoffs.

\section{Pipeline settings}
\label{app:pipeline}

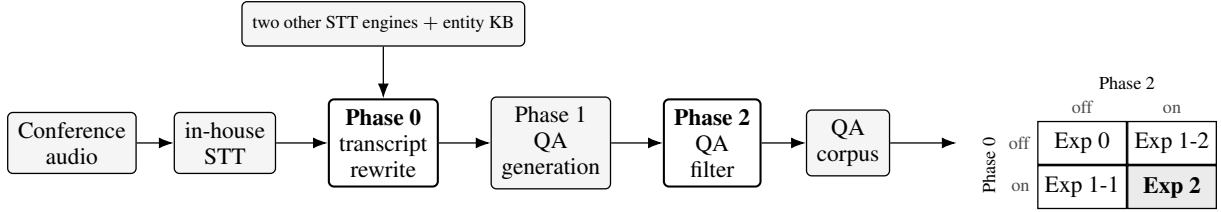
\begin{figure*}[t]
\centering
\resizebox{\textwidth}{!}{%
\begin{tikzpicture}[
  font=\small,
  box/.style={draw, rounded corners=2pt, minimum height=6mm, align=center,
              inner xsep=4pt, line width=0.5pt},
  fixed/.style={box, fill=black!4},
  tog/.style={box, thick, fill=white},
  ar/.style={-{Latex[length=2mm]}, line width=0.5pt},
]

\node[fixed] (audio) {Conference\\audio};
\node[fixed, right=5mm of audio] (stt) {in-house\\STT};
\node[tog,   right=7mm of stt]   (p0)  {\textbf{Phase 0}\\transcript\\rewrite};
\node[fixed, right=7mm of p0]    (p1)  {Phase 1\\QA\\generation};
\node[tog,   right=7mm of p1]    (p2)  {\textbf{Phase 2}\\QA\\filter};
\node[fixed, right=6mm of p2]    (out) {QA\\corpus};

\foreach \a/\b in {audio/stt, stt/p0, p0/p1, p1/p2, p2/out}
  \draw[ar] (\a) -- (\b);

\node[fixed, above=6mm of p0, yshift=1mm] (evid)
  {\scriptsize two other STT engines $+$ entity KB};
\draw[ar] (evid) -- (p0);

\begin{scope}[shift={([xshift=26mm]out.east)}]
  \tikzset{
    cell/.style={draw, minimum width=12mm, minimum height=5.8mm,
                 inner sep=1pt, font=\small, line width=0.5pt},
    hdr/.style={font=\scriptsize, inner sep=1pt, text=black!70},
  }
  \node[cell]                                 (c00) at (0,0) {Exp 0};
  \node[cell, right=0pt of c00]               (c01)          {Exp 1-2};
  \node[cell, below=0pt of c00]               (c10)          {Exp 1-1};
  \node[cell, below=0pt of c01, fill=black!8] (c11)          {\textbf{Exp 2}};

  \node[hdr, above=0.6mm of c00] {off};
  \node[hdr, above=0.6mm of c01] {on};
  \node[font=\scriptsize] at ($(c00.north)!0.5!(c01.north)+(0,5.2mm)$) {Phase 2};

  \node[hdr, left=0.6mm of c00] {off};
  \node[hdr, left=0.6mm of c10] {on};
  \node[font=\scriptsize, rotate=90]
    at ($(c00.west)!0.5!(c10.west)+(-6.5mm,0)$) {Phase 0};
\end{scope}
\draw[ar] (out.east) -- ([xshift=9mm]out.east);

\end{tikzpicture}%
}
\caption{The pipeline and the four ablation conditions. Bold boxes are the two
stages we switch on and off.}
\label{fig:pipeline}
\end{figure*}

\paragraph{Phase 0.} The rewriting model reads five transcript sentences at a
time at temperature $0$. It receives the in-house transcript, two transcripts of
the same audio from commercial ASR models, and entity candidates retrieved from a
domain knowledge base with $k=3$ and a similarity floor of $0.15$. Entity
extraction runs on a separate small model. When a batch fails, the pipeline keeps
the original sentences and continues.

\paragraph{Phase 2.} The judge is a single model, \texttt{gpt-5.2}, at
temperature $0$. It scores four dimensions from $0$ to $10$. A pair passes when
every dimension reaches $7.0$ and fails when any falls below $4.0$. In between it
is marked for human review. Every pair that does not pass, failed or marked alike,
enters a refinement loop of at most three rounds, which stops early when the pair
passes, when the score drops, or when the gain falls below $0.5$. The export keeps
passed and refined pairs and drops the rest, so no pair waits on a human. Deduplication uses embedding similarity, at $0.95$ in the
early pass and $0.85$ in the late one.

\begin{table}[t]
\centering\small
\setlength{\tabcolsep}{4pt}
\begin{tabular}{lrr}
\toprule
Fate of an input pair & 0 $\to$ 1-2 & 1-1 $\to$ 2 \\
\midrule
Unchanged, character for character & \textbf{67.8\%} & \textbf{64.2\%} \\
Kept, answer reworded    & 0.7\% & 1.0\% \\
Kept, answer expanded    & 0.2\% & 0.4\% \\
Kept, a fact changed     & 0.0\% & 0.1\% \\
Dropped or rewritten     & 31.3\% & 34.3\% \\
\bottomrule
\end{tabular}
\caption{What Phase~2 does to the pairs entering it, matched on exact question
text. Columns name the two contrasts that toggle Phase~2 alone.}
\label{tab:phase2}
\end{table}

\paragraph{Cost accounting.} We price recorded calls from their prompt and
completion tokens at the provider's public per-token rates. Per generated pair,
generation costs about \$$0.016$, Phase~0 about \$$0.017$, and the Phase~2 judge
about \$$0.006$. Phase~0 reads roughly six times the text Phase~2 sees, so
turning it on doubles what a pair costs.

\paragraph{Sessions.} Medical sessions cover surgery, cardiology, endocrinology,
dentistry, and oncology. Finance sessions cover venture capital, banking,
fintech, AI policy, and trade.

\section{Evaluation settings}
\label{app:setup}

\paragraph{Corpus quality.} Four LLM judges score the 200-QA sample. The
frontier-tier pair is \texttt{opus-4.7} and \texttt{gpt-5.4}, and the
production-tier pair is \texttt{sonnet-4.6} and \texttt{gpt-4o}. Two come from
each provider, so no single vendor decides the ranking. Six human domain experts, three per
domain, score the same sample. Every rater sees inputs with no condition label
and no source label. The frontier-tier pair is a subset of the four, not an
independent third source; we report it separately only to show that the ranking
survives when the weaker two judges are dropped.
Table~\ref{tab:rubric} gives the rubric every rater follows.

\begin{table}[t]
\centering\small
\setlength{\tabcolsep}{4pt}
\begin{tabular}{@{}lp{5.1cm}@{}}
\toprule
Dimension & What the rater judges \\
\midrule
Faithfulness & The answer is factually right and invents nothing. \\
Domain accuracy & The medical or finance content is right. \\
Question quality & The question is worth learning from. Plain recall scores $1$--$2$, reasoning or synthesis $4$--$5$. \\
Answer depth & The answer lists facts ($1$--$2$) or reasons through cause, comparison, or synthesis ($4$--$5$). \\
Coherence & The answer addresses the question it was asked. \\
\bottomrule
\end{tabular}
\caption{The quality rubric. Each dimension runs $1$ to $5$, where $1$ is very
poor, $3$ acceptable, and $5$ excellent. Raters are told not to reward length,
since a long recall answer is still shallow. A sixth dimension checks whether the
difficulty and type labels match the content. It rates the labels rather than the
pair, so it stays out of the average.}
\label{tab:rubric}
\end{table}

\paragraph{Retrieval.} We run EXAONE-3.5 at 2.4B and 7.8B, Gemma-3 at 4B,
Llama-3.2 at 3B, and Qwen3.5 at 4B and 9B, all instruction-tuned and
open-weight. Retrieval uses
\texttt{bge-m3} with $k=3$. Generation is greedy with a 512-token limit. Six
models and two domains give 12 cells. The four ablation conditions plus two
controls run on all 163 questions. A domain expert wrote 43 of those, and the
two Whisper conditions run on that subset, of which 33 pass the wording filter. That is $6{,}384$ generated answers, each scored by both
judges. Headline retrieval results are computed on the 115 questions that pass
the wording filter of \S\ref{sec:setup:retrieval}.

\paragraph{Judges.} Correctness judging uses \texttt{gpt-5.6-sol} and
\texttt{claude-sonnet-5}, and we average their scores.

\paragraph{Novelty.} Both models answer every question closed-book, and each
grades the other's answers on \emph{novel}. Neither grades its own output. The
\emph{grounded} and \emph{correct} axes judge our own pair rather than the
closed-book answer, so both models grade every pair on those, and a pair passes
only when both say yes. A pair counts as effectively novel when it clears those
two axes and the novel axis, under whichever of the two counts is in use. Closed-book answers depend only on the question, so we ask each of the
$6{,}886$ distinct questions once and map the answer back to every condition
that uses it, giving $10{,}695$ graded pairs. A question that repeats inside its
own condition is graded once, so the graded set is two questions and three pairs
smaller than the corpora.

\section{Why novelty had little room}
\label{app:novelty}

Table~\ref{tab:axes} reports each frontier model separately and both counts.
Only the strict count appears in the body, since the two gating axes barely move
and the two models agree closely.

\begin{table}[t]
\centering\small
\begin{tabular}{lrrrr}
\toprule
Missed by & Exp~0 & Exp~1-1 & Exp~1-2 & Exp~2 \\
\midrule
\texttt{sonnet-5}  & 45.0 & 44.3 & 46.0 & 44.6 \\
\texttt{gpt-5.6}   & 46.1 & 45.9 & 47.1 & 47.3 \\
\midrule
either model       & 55.6 & 54.2 & 56.1 & 55.3 \\
both models        & 35.6 & 36.1 & 36.9 & 36.5 \\
\bottomrule
\end{tabular}
\caption{Percent of pairs whose core content the closed-book answer misses.
Model names abbreviate \texttt{claude-sonnet-5} and \texttt{gpt-5.6-sol}. Every
row already requires the grounded and correct axes, and a pair clears those only
when both models say yes. The last two rows are the lenient and strict
counts of \S\ref{sec:setup:novelty}.}
\label{tab:axes}
\end{table}

We built this measure expecting Exp~0 to carry transcription errors and filler,
which appear in no reference and so count as novel. That is not what happened.
The grounding axis stays near $99.6\%$ everywhere, and it is a sensitive
check rather than one that always passes (App.~\ref{app:confound}). There was
almost no noise to filter, because Exp~0 is not a raw transcript. Phase~1 runs
in every condition, so even Exp~0 is QA that a language model wrote from the
transcript. Filler and recognition errors rarely survive that step. The starting
point was clean, so the two stages had little room to help.

\section{Question sources}
\label{app:qsource}

The wording filter of \S\ref{sec:setup:retrieval} works per question. We take
the gap in character 4-gram overlap between the reference answer and the
Exp~2 and Exp~0 pairs from that question's session, and we keep the question
when that gap is at most $10.9$ percentage points in either direction. The threshold is the 80th
percentile of the same quantity on the expert-written questions, computed
without looking at any accuracy score. It keeps 33 of the 43 expert questions
and 82 of the 120 model-written ones.

We also check the outcome rather than trusting the protocol. First, the
transcripts the model read are the corrected ones, not the raw
ASR output. They therefore cannot sit closer to the unrefined corpora, though
they could sit closer to the refined ones. Second, we measure how much of each
corpus appears verbatim in those transcripts. The four corpora sit within $2.3$
points of one another, at $28.0$, $28.9$, $26.6$, and $27.7$ percent for Exp~0,
Exp~1-1, Exp~1-2, and Exp~2, so the lean the first point allows does not
appear. Third, we compute session-level recall separately for the two question
sets. The Exp~0 minus Exp~2 gap is $+0.0$ points on the expert questions and
$+2.5$ on the model-written ones, and the difference between those gaps is
$[-4.7, +9.5]$, which covers zero. No condition is favored by where the
questions came from.

\paragraph{Sensitivity to the wording threshold.} The threshold decides which
questions we keep. For each question we measure the gap in character
4-gram overlap between its reference answer and the Exp~2 and Exp~0 pairs from
the same session, and we keep the question when that gap stays under the
threshold in either direction. A low threshold keeps only questions whose wording leans toward
neither corpus, at the cost of a smaller set. We fixed ours at $10.9$\,pp from
the expert questions without looking at any accuracy score, but a reader may ask
what other choices would give. Table~\ref{tab:sens} sweeps it.

\begin{table}[t]
\centering\small
\setlength{\tabcolsep}{4pt}
\begin{tabular}{lrrrc}
\toprule
Threshold & Qs & 1-1 $-$ 0 & 2 $-$ 0 & Equiv. \\
\midrule
$4$\,pp    &  52 & $-0.040^{*}$ & $-0.029^{*}$ & no  \\
$6$\,pp    &  68 & $-0.058^{*}$ & $-0.023^{*}$ & yes \\
$8$\,pp    &  95 & $-0.070^{*}$ & $-0.031^{*}$ & no  \\
\textbf{$10.9$\,pp} & \textbf{115} & $\mathbf{-0.061^{*}}$ & $\mathbf{-0.020}$ & \textbf{yes} \\
$14$\,pp   & 125 & $-0.050^{*}$ & $-0.014$     & yes \\
$18$\,pp   & 138 & $-0.030$     & $+0.005$     & yes \\
$25$\,pp   & 148 & $-0.022$     & $+0.009$     & yes \\
$40$\,pp   & 158 & $-0.006$     & $+0.031^{*}$ & no  \\
no filter  & 163 & $-0.007$     & $+0.023^{*}$ & yes \\
\bottomrule
\end{tabular}
\caption{Both ablation contrasts as the wording threshold varies. Columns are
the threshold, the number of questions kept out of 163, the Exp~1-1 and Exp~2
contrasts against Exp~0, and whether the $90\%$ interval for Exp~2 falls inside
the $\pm 0.045$ bound we registered in advance (App.~\ref{app:stats}). An asterisk marks a $95\%$ interval that excludes zero. The bold row is the one we use.}
\label{tab:sens}
\end{table}

Both estimates move toward positive as the threshold admits more pairs whose
reference answers lean toward the refined corpora, which is the direction the
wording effect predicts. The two contrasts behave differently under that
pressure. Exp~2 against Exp~0 changes sign, while the rewrite penalty stays
negative at every threshold and remains significant at the five tightest. No
estimate of the combined contrast exceeds $0.031$ in size against a bound of
$0.045$, and equivalence holds at six of the nine settings, including no filter
at all.

\paragraph{Session-level recall.} Each question was written from one session's
transcript, so the pairs that can answer it are the pairs generated from that
session. We score a question as a hit when at least one of the top three
retrieved pairs comes from its own session. We use the session rather than the
individual pair because the four corpora share almost no pairs once
Phase~0 is on, which leaves the session as the only coordinate that survives
every condition. Provenance comes from the directory layout of the generation
output, and it resolves $99.7$ to $100\%$ of corpus pairs. The retriever,
the index, and $k$ match the generation experiment.

\paragraph{Expert agreement.} All six domain experts rank Exp~2 above Exp~0. A
sign test on six independent raters gives $p = 0.031$ two-sided, so the
shared direction is unlikely to arise by chance.

\paragraph{Phase~0 coverage.} Phase~0 runs in batches, and a failed batch
would leave the original sentences in place and weaken the treatment. Across the
61 runs with recorded metadata, all $5{,}182$ batches returned a response, and
every run logged processing for the full transcript, a median of 419 sentences.
Phase~0 covered every session it was applied to.

\paragraph{Judge agreement.} On the $4{,}536$ answers behind the main analysis
the two correctness judges agree exactly $82\%$ of the time. Cohen's $\kappa$ is
$0.726$, and squared-weighted $\kappa$, which counts a one-step disagreement as
smaller than a two-step one, is $0.875$. Eight answers are called
\emph{correct} by one judge and \emph{incorrect} by the other. The main contrast
holds under either judge alone. Exp~2 minus Exp~0 is $-0.019$ with
\texttt{gpt-5.6-sol} and $-0.020$ with \texttt{claude-sonnet-5}, against the
$-0.020$ we report from their average.

\paragraph{Validating the correctness axis.} Effective novelty requires a pair
to pass a \emph{correct} check, so we test that check against independent
raters. We draw 120 pairs on which the two judges agreed, 60 called correct and
60 called incorrect, and give them to two raters who see only the question and
the answer, with no judge verdict and no condition label. Of the 60 pairs the
judges called correct, the raters confirm 60 and 59. Of the 60 called
incorrect, they confirm 39 and 47, so the judges reject somewhat more than
independent raters would. The correct axis removes $1{,}029$ of the $10{,}695$
pairs, so this over-rejection reaches a few percent of the set at most, and it
pushes effective novelty down rather than up. The two raters agree with each
other on $92.5\%$ of the sample.

\section{Statistical methods}
\label{app:stats}

\begin{table*}[t]
\centering\small
\begin{tabular}{lrr}
\toprule
Contrast & Cell means & Mixed effects \\
\midrule
Exp~1-1 vs.\ Exp~0 & $-0.061\ [-0.090, -0.032]$ & $-0.058\ [-0.083, -0.034]$ \\
Exp~1-2 vs.\ Exp~0 & $+0.018\ [+0.004, +0.031]$ & $+0.019\ [-0.005, +0.043]$ \\
Exp~2 vs.\ Exp~0   & $-0.020\ [-0.042, +0.003]$ & $-0.018\ [-0.042, +0.006]$ \\
\bottomrule
\end{tabular}
\caption{The three ablation contrasts under both analyses, with $95\%$
intervals. Cell means resample the 12 model-by-domain cells. The mixed-effects
fit carries random effects for session, question, and model over the
$2{,}760$ scored answers.}
\label{tab:mixed}
\end{table*}

Retrieval contrasts are paired at the question level, since the two sides
answer the same evaluation questions. We take the mean paired difference inside each model-by-domain
cell, and the percentile bootstrap resamples those 12 cell means, with
$10{,}000$ draws and a fixed seed. Contrasts between disjoint pair sets, such as
the non-shared Phase~2 comparison, cannot be paired and instead bootstrap the
difference of two independent means. Because we pair before averaging, a contrast
need not equal the difference of the two rounded means in Table~\ref{tab:main}.

We report nine planned contrasts: three ablation contrasts against Exp~0
(Exp~1-1, Exp~1-2, Exp~2), two anchoring contrasts (retrieval against no
retrieval, and matched against wrong-domain retrieval), two transcript-source
contrasts (the two Whisper conditions against their in-house counterparts), and
two Phase~2 contrasts on novelty (Exp~0 against Exp~1-2, Exp~1-1 against
Exp~2). Under Bonferroni correction the threshold is $0.05/9 = 0.0056$.
Table~\ref{tab:contrasts} lists every one with its $p$ value; it is the single
place we state significance, and the rest of the paper follows it.

\paragraph{Where the equivalence bound comes from.} An equivalence test needs a
smallest effect worth caring about, and choosing one after seeing the result
would let us reach whichever conclusion we prefer. We therefore derived ours from
our own quality measurement and registered it before running the experiment. The
judges score on a $1$--$5$ scale, whose range is 4 points, and refinement moved
them by $+0.18$ to $+0.22$. Retrieval accuracy runs on a $0$--$1$ scale, whose
range is 1 point. A quality gain of $+0.18$ that carried to retrieval one for one
would therefore show up as $0.18/4 = 0.045$ in accuracy. We take the low end of
the measured range rather than the high one, since a larger bound would make
equivalence easier to declare. The test is a TOST, or two one-sided tests, on the
Exp~2 against Exp~0 contrast over the 12 cells, with bounds at $\pm 0.045$ and one-sided $\alpha =
0.05$ on each side, and it declares equivalence when the $90\%$ interval lies
inside those bounds.

\paragraph{Do the two stages add up?} Taken alone the rewrite gives $-0.061$ and the
filter $+0.018$. Those would sum to $-0.043$, while running both gives
$-0.020$. The interaction contrast, Exp~2 minus Exp~1-1 minus Exp~1-2 plus
Exp~0, is $+0.024$ $[-0.006, +0.053]$, $p=0.14$. We cannot reject additivity.

\paragraph{Mixed-effects check.} Cell means treat the twelve cells as
independent, but the same questions appear in every cell and the questions sit
inside sessions. We refit the three ablation contrasts over the $2{,}760$ scored
answers of the main analysis, with random effects for session, question, and
model. Each coefficient is close to the cell summary it replaces
(Table~\ref{tab:mixed}). Only Exp~1-2 differs in whether its interval covers
zero. It excludes zero in the cell summary and covers it here. The body already reports that gain as
nominal. Model identity carries the largest
random-effect variance at $0.049$, against $0.008$ for session and $0.002$ for
question. Models differ in overall accuracy rather than in how they respond to
the two stages.

We decompose the variance of the Exp~2 minus Exp~0 cell means into a
between-cell and a within-cell part. The between-cell part is zero. The
observed variance of cell means, $0.0029$, is smaller than the within-cell
variance divided by the number of questions. The twelve cells are therefore not
distinguishable from one another, and the clustering that matters is at the
question and session level, not the model level.

\section{Non-shared Phase 2 comparison}
\label{app:nonshared}

Phase~2 leaves most pairs untouched, so a contrast over all pairs mixes those
that can differ with those that cannot. Restricting to pairs that appear in only
one side of a contrast isolates what Phase~2 actually changed
(Table~\ref{tab:nonshared}).

\begin{table}[t]
\centering\small
\begin{tabular}{lrrr}
\toprule
Contrast & Shared & All pairs & Non-shared \\
\midrule
Exp~0 $\to$ Exp~1-2 & $69\%$ & $+1.4$ & $+4.2$ \\
Exp~1-1 $\to$ Exp~2 & $66\%$ & $+0.4$ & $+2.2$ \\
\bottomrule
\end{tabular}
\caption{Effective novelty under the strict count, in percentage points. Shared
is the share of pairs whose question appears on both sides, which runs slightly
above the share left unchanged character for character
(Table~\ref{tab:phase2}). The last two columns score all pairs and then only the
pairs one side does not share.}
\label{tab:nonshared}
\end{table}

The non-shared intervals are $[-0.4, +8.8]$ and $[-2.4, +6.9]$, both covering zero.
On shared pairs the contrast is $+0.02$ and $+0.04$ points, confirming that
identical text receives near-identical judgments.

\section{Confound checks}
\label{app:confound}

\paragraph{Lexical drift under Phase~0.}
Phase~0 rewrites the transcript, so the corpus wording moves away from the
wording of an incoming question. Retrieval could then fail for reasons unrelated
to content quality. We measure the drift as character 3-gram overlap between each
question and the three passages it retrieves, averaged over the three. Overlap
falls from $0.0535$ in Exp~0 to $0.0512$ in Exp~1-1, $0.0503$ in Exp~1-2, and
$0.0480$ in Exp~2. The drops for Exp~1-2 and Exp~2 both exclude zero.

The drift is real but too weak to carry the result. Across pairs, the change in
overlap correlates with the change in accuracy at $r = 0.10$, which explains
about one percent of the variance. Regressing accuracy change on overlap change
and reading the intercept gives $-0.031$ ($t=-2.42$) for Phase~0 alone,
$+0.025$ ($t=+2.39$) for Phase~2 alone, and $-0.013$ (not significant) for both.
Both single stages keep their sign after the confound is controlled.

\paragraph{Transcript quality interaction.}
The gap between the in-house transcript and Whisper-medium is $+0.009$
$[-0.016, +0.035]$ with both stages off and $-0.001$ $[-0.045, +0.042]$ with both on.
Testing the difference between the two gaps on the cell unit gives $+0.011$
$[-0.026, +0.048]$ with $t = 0.57$. Detecting a change of this size would need
more than 12 cells.

The wording effect reaches this contrast too. The reference answers come from
the corrected in-house transcript, so a question that leans toward the refined
corpora leans toward the in-house ones for the same reason. Scored on all 43
expert questions instead of the 33 that pass the wording filter, the stages-off
gap reads $+0.045$ $[+0.020, +0.070]$. Removing those ten questions removes most
of the in-house advantage.

\paragraph{Does the grounding axis fire?}
A dimension that returns true on $99.6\%$ of pairs could be measuring nothing. It
is not. The two judges flag 35 and 13 pairs, and 5 of those overlap, against $0.04$
expected by chance. The flagged text is genuine transcription damage, not
arbitrary text. The axis is sensitive; the signal is small because there is
little noise to find.

\section{Additional observations}
\label{app:extra}

Judges can reward a longer answer, so the gain from supplying a corpus could be
an artifact of answer length. It is not. Answers written with the corpus are
shorter than closed-book ones, $272$ characters against $346$, and they score
higher.

\paragraph{How models fail under wrong-domain retrieval.}
Models fail in two ways. Four of the six stop answering and score near zero,
$0.030$ and $0.000$ for Qwen at 4B and 9B, $0.059$ for Gemma and $0.068$ for
Llama. EXAONE keeps answering and lands near its own closed-book baseline,
$0.225$ against $0.229$ at 2.4B and $0.218$ against $0.280$ at 7.8B. Every model
scores at or below its baseline, so the drop is not one family's artifact.

\begin{table}[t]
\centering\small
\begin{tabular}{lrrrr}
\toprule
Source & Exp~0 & Exp~1-1 & Exp~1-2 & Exp~2 \\
\midrule
4 LLM judges      & 3.81 & 3.90 & 3.84 & \textbf{3.99} \\
6 human experts   & 3.36 & 3.46 & 3.46 & \textbf{3.58} \\
\midrule
2 frontier judges & 3.41 & 3.50 & 3.45 & \textbf{3.65} \\
\bottomrule
\end{tabular}
\caption{The first two rows repeat Table~\ref{tab:main}. The third is the
frontier-tier subset of the LLM panel, which shows the ranking without the
weaker two judges.}
\label{tab:quality}
\end{table}

\paragraph{Per-dimension quality scores.}
All five rated dimensions rise from Exp~0 to Exp~2: faithfulness $3.89$ to
$4.06$, domain accuracy $3.88$ to $4.05$, question quality $3.42$ to $3.60$,
answer depth $3.15$ to $3.40$, and coherence $4.70$ to $4.84$.

\section{Secondary results}
\label{app:secondary}

\paragraph{Transcript quality.}
Two of the eight retrieval conditions use corpora built from Whisper-medium
transcripts instead of the in-house transcript. Neither gap is detectable on the
33 questions that pass the wording filter. The gap is $+0.009$ $[-0.016, +0.035]$ with both
stages off and $-0.001$ $[-0.045, +0.042]$ with both on, each positive in 5 of 12
cells. Appendix~\ref{app:confound} gives the interaction test and the unfiltered
counterpart.

\paragraph{Fine-tuning.}
We fine-tuned two model families on the four corpora with a single standard
recipe and two seeds each, and evaluated on multiple-choice benchmarks. Fine-tuning lowers accuracy
by $1.6$ points when each model is measured at the prompt format that suits it
best, and only 9 of 96 model-task cells land above the un-fine-tuned model.
Fine-tuning aligns a model to a chat format, so scoring both sides under raw
completion penalizes the fine-tuned one. Measuring every model in one fixed
format that way reports a $3.1$-point drop, so part of the apparent damage is a
format mismatch rather than a loss of ability. The sign does not change either
way. The stage contrast, Exp~2 against Exp~0, is $+0.51$ percentage
points and is not significant once the unit of analysis rises above the
individual cell. This experiment covers two models rather than the full set, so we report it as a consistency check with the other two instruments rather
than as an independent result.

A drop on every corpus invites the reading that the training recipe, not the
data, is at fault. Three signatures would show that, and none appears. Forgetting
would hit the general benchmark hardest, but MMLU falls least, $-0.98$ points
against $-1.93$ on the domain benchmarks. Overfitting would scale the damage with
training set size, but the correlation across conditions is $-0.27$ and runs the
wrong way. Instability would leave seed-to-seed spread larger than the effect,
but the median spread is $0.46$ points against the $1.6$-point drop. The drop is
also small in absolute terms. MMLU moves from $71.8\%$ to $70.8\%$. We fit LoRA
adapters at rank 16 with a learning rate of $2 \times 10^{-4}$ for three epochs,
which is a conventional setting and adapts far less of the model than full
fine-tuning. The instrument works; it simply finds no gain.

\end{document}